%% file: main.tex
\documentclass{article}
\usepackage{ital-ia}
\usepackage[utf8]{inputenc}
\usepackage[T1]{fontenc}

\usepackage{times}
\usepackage{soul}
\usepackage{url}
\usepackage[hidelinks]{hyperref}
\usepackage[utf8]{inputenc}
\usepackage[small]{caption}
\usepackage{graphicx}
\usepackage{amsmath}
\usepackage{booktabs}
\usepackage{algorithm}
\usepackage{algorithmic}
\usepackage{fancyvrb}
\usepackage{multirow}
\usepackage{tabularx}

\title{Information Extraction through AI techniques: The KIDs use case at CONSOB}

\author{
	Domenico Lembo$^1$ \and
	Alessandra Limosani$^2$ \and
	Francesca Medda$^2$ \and \\
	Alessandra Monaco$^{1}$\and
	Federico Maria Scafoglieri$^1$
	\affiliations
	$^1$Sapienza Università di Roma\\
	$^2$Commissione Nazionale per le Societ\`a e la Borsa 
	\emails
	$^1$\{lembo, monaco, scafoglieri\}@diag.uniroma1.it,
	$^2$\{a.limosani,f.medda\}@consob.it
}

\input{macros/macros}

\begin{document}

\maketitle

\input{00-Abstract/abstract}

\input{01-Introduction/introduction}
\input{02-UseCase/usecase}

\input{03-Approach/approach}

\input{04-Results/results}
\input{05-Conclusion/conclusion}

\bibliographystyle{abbrv}
\bibliography{ital-ia}

\end{document}

%% file: macros/macros.tex
\newcommand{\Consob}{Consob}

%% file: 00-Abstract/abstract.tex
\begin{abstract}
		In this paper we report on the initial activities carried out within a collaboration between \Consob\ and Sapienza University.
		We focus %specifically 
		on Information Extraction %(IE)
		from documents describing financial instruments. %, like bonds or insurance contracts. %Packaged Retail Investment and Insurance-based Investments Products (PRIIPs). 
		%The project is carried out within a program through which CONSOB is establishing a stable collaboration with various Italian Universities. %Sapienza being one of the first involved in this program. 
%		We discuss our approach, which is based on the idea of structuring extracted data in the form of a Knowledge Graph, and reports on two different implementations, aiming at pursuing an efficient, easy-to manage semantic IE approach.
We discuss how we automate this task, via both rule-based and machine learning-based methods, %based on a combination of rule-based and machine learning-based techniques, 
and provide 
our first results.
\end{abstract}

%% file: 01-Introduction/introduction.tex
\section{Introduction}

%, where the excepted output was the creation of a rich knowledge graph starting from financial documents.
%In Italy, supervising the financial market to detect illicit activities is entrusted to CONSOB that daily receives from the creators of financial instruments (a.k.a. financial manufacturers) helpful documentation, namely Key Information Documents (KIDs), that should validate the non-existence of malfeasance regarding financial products. On this, CONSOB carries out a sample-based mainly manually activity of searching for certain specific data in the documents.
%%

%In 2019 \Consob\ (Commissione Nazionale per la Società e la Borsa) initiated a collaboration program with some Italian universities, aimed at financing research activities  and promoting technological transfer of the main achieved results in the \Consob\ business domain. 

In Italy, \emph{Commissione Nazionale per le Società e la Borsa (Consob)} is the supervisory and regulatory authority of the financial market. Among the several functions, Consob has the role of monitoring and supervising any financial instruments that are issued, with the ultimate aim of detecting and enforcing against illicit conduct. In this project, Consob, rather than outsourcing the digital activity to a specific operator, has taken the decision to develop the project through a collaboration between expertise from Academia and its own expertise. 

The objective of such an approach is twofold: first to develop a bottom-up approach driven by the actual needs of supervisory activity; and second, to explore and experiment in innovative ways the improvement of supervisory effectiveness. The results to date show not only the digital improvement but also a direct involvement and participation of the Consob teams and thus effective knowledge spill-overs from one Division to another have taken place.

The specific project we will discuss in the following, besides \Consob, involves Sapienza University of Rome, and namely a team from the department of  Computer, Control, and Management Engineering, with expertise in 
%has been selected within the above mentioned program for its competences in the area of 
Artificial Intelligence and Information Management. 
%%
%The first project developed as part of this collaboration stems from a specific demand of \Consob\ concerning with the  need to support, and speed up, the process of verification of some text documents describing financial products. This issue is particularly critical for \Consob\ for the large amount of documents to be processed and the importance of the data they contain, which can be exploited for useful analysis.
%%
To support \Consob\ in its supervision activities, and in particular in the task of verifying the correctness and completeness of the information about financial instruments, within this collaboration a solution for the automatic extraction of structured information from free text documents is being developed. Through such solution, relevant data contained in the documents are not to be manually identified and extracted, but can be collected through an automated process which make them available in a machine processable format.
%, thus strongly speeding up the surveillance activity of \Consob\ over the documents and make it possible the development of advanced data analysis process.
%%
To this aim, essentially two approaches to automatic Information Extraction (IE) have been adopted: rule-based and machine learning (ML)-based.
In the rest of the paper we describe them, explain why both are useful to achieve our goals, and report on the first results we obtained in the project, which have been partially reported, limited to the rule-based approach, in~\cite{scafoglieri2021boosting}.

\begin{figure}[h]
	\begin{center}
		\vspace*{-0.18in}
		\includegraphics[width=0.47\textwidth]{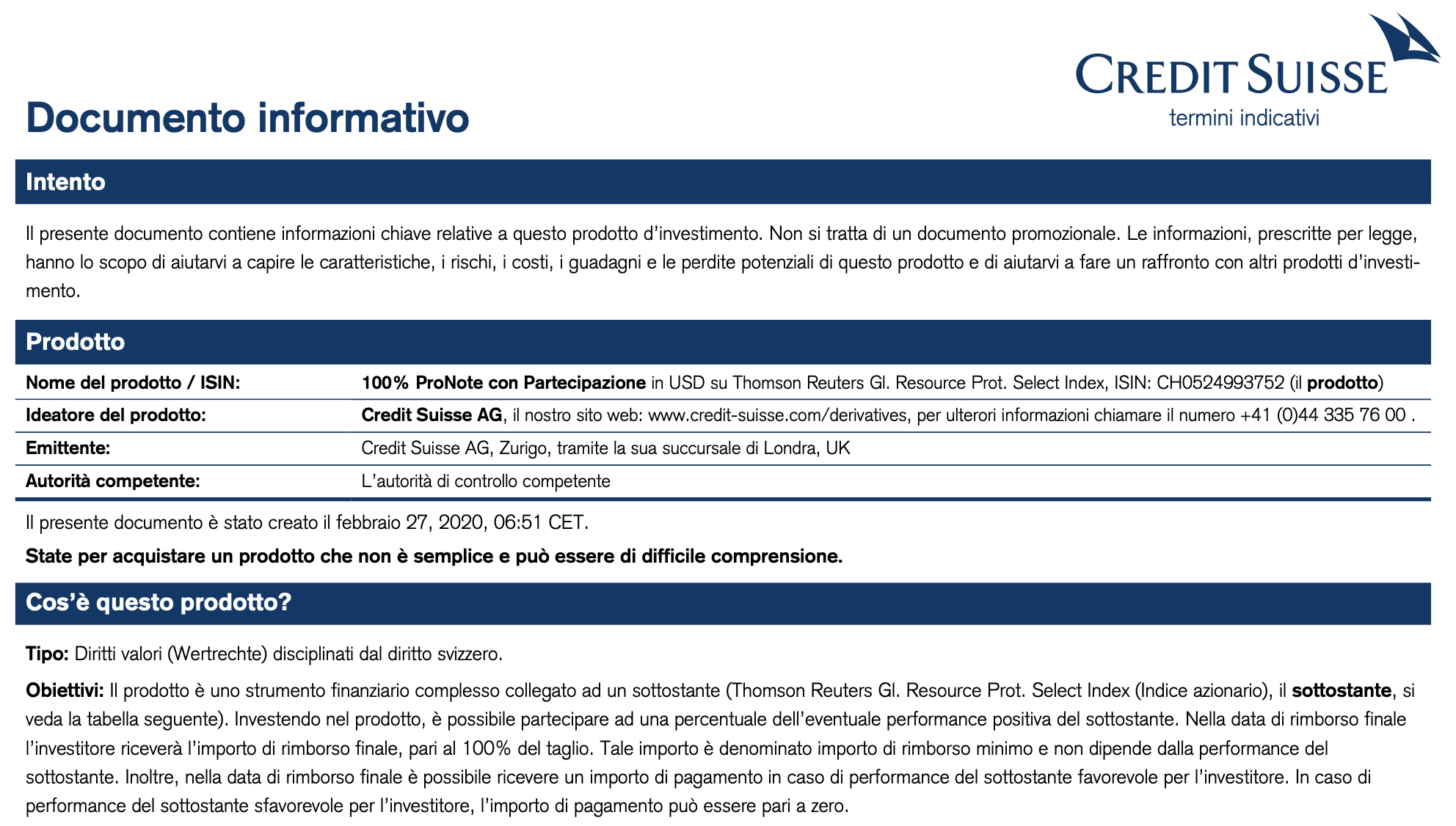}
		%\caption{KID by CreditSuisse. The full document can be found in Appendix}
		\vspace*{-0.1in}
		\caption{Portion of a KID by CreditSuisse.}
		\label{fig:KID_1}
	\end{center}
\end{figure}
\vspace*{-0.3in}

%% file: 02-UseCase/usecase.tex
\section{The use case}
In the EU, the creators of financial products (a.k.a. financial manufacturers) are obliged by law\footnote{PRIIPs Regulation n. 1286/2014} to make information related to so-called PRIIPs (Packaged Retail Investment and Insurance-based Investments Products) publicly available. The NCAs (National Competent Authorities) have %indeed 
supervisory duties on such products, so that they can be safely placed on the respective 
national markets. 
The legislation requires information about PRIIPs to be communicated to NCAs through documents called KIDs (Key Information Documents). In the practice, this means that features to be checked are cast into text reports, typically formatted as pdf files (cf.\ Figure~\ref{fig:KID_1}), and extracting structured data from them (to bootstrap control activities), is actually in charge to the authority (In Italy, CONSOB). Due to the massive amount of documents to be analyzed (e.g., $\sim$700.000 KIDs received by CONSOB in 2019, more than 1 million in 2020), %of course 
this process cannot be carried out manually, but still it is only partially automated to date.
%%%%%%%%%%%%%%%%%%%%%%%%%%%%%%%%%%%%%%%%%%%%%%%%%%%%%%%%%%%%%%%%%%%%%%%%%%%%%

%% file: 03-Approach/approach.tex
\section{The approach}

%descrivere l'architettura/pipeline e motivare l'approccio a regole (alta precisione e recall; dominio in cui sono disponibili regole di estrazione semplicemente da codificare). Discutere il problema dell'estrazione da tabelle. E quindi utilizzo di un approccio ML

The KIDs present key information about financial products through free text or tables.
We decided to extract relevant data contained in the free text part of the documents through a rule-based mechanism using language that is an evolution of regular expressions~\cite{doleschal2021database,lembo2020ontology}. This choice lies in the fact that the KIDs follow a quite rigid template imposed by the European authority, a characteristic that makes domain experts able to express quite precise patterns for the identification of wanted data, and AI experts to encode them into extraction rules.
%easier to manual coding of these rules also thanks to the support of domain experts. 

%This approach has guaranteed, as we will show in the following, an high accuracy and recall without losing the explainability charateristic of rule-based systems.

%As for the information contained in the free text, it is extracted by means of a rule-based language, the evolution of regular expressions. These rules have been coded manually with the support of experts in order to guarantee high accuracy.

%For these reasons, the approach chosen to extract information from them is a mix of manually written rules and learning-based systems. The rules, written manually through an advanced language based on regular expressions, work on the free text portions extracted from the pdf and identify with high accuracy the information searched. 

%As far as tables are concerned
However, rule-based mechanisms do not lend themselves well to the extraction of data contained in tables, which are particularly difficult to identify in pdf documents, depending also on the way in which the table is originally formatted. For this reason the chosen extraction approach is based on learning systems that treat the pdf as if it were an image and aim to identify the set of pixels that are tables, reconstructing the cells from which the information sought will be identified. In the following we will discuss in more detail the two approaches highlighting their complete pipelines and showing the results obtained.

\subsection{Rule-based IE}

The rules we defined deal with the extraction of 16 information fields from the free text contained in a KID. Among them we mention the product name, the product ID (a.k.a. ISIN), the manufacturer, the investment risk, and so on. 

The rule-based extraction, is organized in three macro modules described in the following, each of them dealing with a specific task.

%%SI PUO'LEVARE
%\includegraphics[width=0.9\columnwidth]{03-Approach/Rulepipeline}

%\begin{itemize}
%	\item \textbf{Data Preparation}: The KID in PDF format is transformed into pieces of annotable text.
%	\item \textbf{Annotation}: By means of rule-based extractors, the annotations useful to provide the final output are generated.
%	\item \textbf{Export}: The output of the tool is arranged in the form of a CSV file.
%\end{itemize}
\vspace{0.05cm}
\noindent
\textbf{Data Preparation:}
The Data Preparation Module transforms the PDF into plain text and clean it from errors.
This step is very critical, and if the PDF-to-text transformation  produces bad quality results, many errors do actually occur in the subsequent phases. 
%(these errors are indeed caused by malformed sentences in the resulting text documents, rather than by the inability of the tool to process the language)
For this reason, we performed it using PDFBox\footnote{\url{https://pdfbox.apache.org}} by Apache, a state-of-the-art and highly customizable library for working with PDF documents. PDFBox presents a large number of options to manage such transformation,  
%These include dealing with different types of characters, normalization of the fonts, managing spaces, reading modes, etc. 
including those for dealing with different types of characters, font normalization, white space management, reading modes, etc.  We set them in order to minimize the number of errors incoming from the transformation.

\vspace{0.05cm}
\noindent \textbf{Annotation:} The annotation task is carried out through CoreNLP, an open-source JAVA toolkit for natural language processing,  whose first release dates back to 2006, and which is still maintained and evolving.
%The principles upon which it is based and which gave rise to its implementation are (\emph{i}) 
%make an 
%easy annotation on text, (\emph{ii}) easy to learn, (\emph{iii}) minimal setup, (\emph{iv}) provide a lightweight framework using plain Java objects, (\emph{v}) transparent communication between components via a common interface.
%Stanford CoreNLP is currently one of the most widely used systems for building text processing tools. Its popularity is due to the fact that in order to use this toolkit, it is necessary to know only JAVA and not specific languages and components management, as it happens, for example, with GATE. Other success factors of CoreNLP have been: complete documentation that favors its immediate use; the software distributed through an open-source license; an active community that has enormously contributed to the project. Although the software was written in JAVA, the community has created several wrappers for other programming languages, including Python, Perl, Ruby, Scala, and Javascript(Node.js).
%Regarding the supported natural languages, the initial version of the software was able to process only English, German, Chinese, Arabic, and French. Over the years, the support has been extended to other languages, mainly by external contributors. For example, as far as Italian is concerned, the extension took place through the CoreNLP-it module~\cite{aprosio2016italy}.
This toolkit provides core NLP services accessible through APIs that can be combined to generate a pipeline for text annotation. Among these services, we mention those related to tokenization, lemmatization, POS tagging, all used in our pipeline.

Another core component of CoreNLP is TokensRegex \cite{chang2014tokensregex}, which extends traditional regular expressions on strings by working on tokens instead of characters and defining pattern matching via a stage-based application. 
These extensions of regular expressions allow for writing %so-called 
\emph{extraction rules}, i.e., rule-based extractors matching on additional token-level features, such as part-of-speech annotations, named entity tags, and custom annotations. 
%In a similar way to the JAPE rules showed in Chapter~\ref{ch:GATE}, these characteristics bring concise rules at a higher level than just matching against the individual words.

We introduce the syntax of a rule written in TokensRegex through the following example and refer the reader to \cite{chang2014tokensregex} for a complete treatment.
Consider the following TokensRegex extraction rule useful for extracting International Securities Identification Numbers (ISINs) from (the free text portion of) KIDs:

{\scriptsize
\begin{Verbatim}
$StartISIN = (
	/ISIN/ /:/ |
    /Codice/ /del/ /Prodotto|prodotto/ /:/ |
	...
)
$EndISIN = (
	/*/
    ...
)
$code = "/([A-Za-z][A-Za-z][0-9]{10})/"
{
    ruleType: "tokens",
    pattern: (
        ($StartISIN) (?$CodeISIN [{word:$code} & 
        {SECTION:"SECTION_PRODUCT"}]+?) ($EndISIN)
        ),
    action: ( Annotate($CodeISIN, ISIN, "ISIN"))
}
\end{Verbatim}
}
%\end{minipage}
\smallskip

The ISIN is an alphanumeric sequence of 12 characters identifying a PRIIP. This sequence begins with two characters, identifying the country code of the product, followed by ten numbers. 
In our TokensRegex rule, the ISIN format is compiled into the %regex
regular expression %identified by 
assigned to $\texttt{\$code}$.
Then, $\texttt{\$StartISIN}$ and $\texttt{\$EndISIN}$ represent the set of sequences of tokens (separated from each other through the symbol \texttt{|}) preceding or following the ISIN code, respectively. 
%The main part of the rule lies in the JSON-like structure at the end of it.
The JSON-like structure at the end of the rule is its main part.
In a nutshell, \texttt{ruleType} specifies how the pattern should be used. In our case, the rule works on tokens specified by \texttt{"tokens"} value. \texttt{pattern} contains the pattern to be matched over the text, built over the tokens using groups like the ones in POSIX (but also with names such as $\texttt{?\$CodeISIN}$). In our example, our pattern specifies how to find the token such that there is a matching with the regex $\texttt{\$code}$ among all the tokens in the product section (i.e. annotated with $\texttt{SECTION\_PRODUCT}$), and such that it is preceded by the tokens in $\texttt{\$StartISN}$ and followed by the ones in $\texttt{\$EndISIN}$. Finally \texttt{action} describes what should happen when the pattern is matched, precisely, what annotation to apply. In our example, we annotate the token identified by the group $\texttt{\$CodeISIN}$ with the annotation $\texttt{ISIN}$.

The result of the application of this rule over the text in Figure~\ref{fig:KID_1} is the annotation of \texttt{CH0524993752} with $\texttt{ISIN}$.

\vspace{0.05cm}
\noindent \textbf{Exporter:} 
%The last task is carried out by the Exporter module. 
The purpose of the Exporter module is to prepare the output in the desired form. Actually, this module is able to transform the text annotations into two possible formats (\emph{i}) CSV (Excel) file or (\emph{ii}) database. %Regarding (\emph{ii}) it opens the doors to arrange data in other formats as well by giving the possibility to add semantics via OBDM approaches \cite{lembo2020ontologyT}.
In particular, option (\emph{ii}) allows to arrange data in various formats, and paves the way for the construction of knowledge graphs on top of the extracted data, e.g., via an OBDM approach~\cite{lembo2020ontologyT}.

%Regarding (\emph{i}), the population of the ontology cannot be done using annotations on the text directly. This is because sometimes some information comes from other sources and cannot be incorporated directly into the rules. 
%Namely, we cannot express directly in TokensRegex rules how to build the URIs denoting certain individuals,
%
%such as URIs relative to the documents, which have to be built  upon their names and file paths (which are not information contained in KIDs).
%This has an obvious impact also in the instantiation of the ontology attributes and relationships that exist between the domain objects obtained by extraction.
%
%To overcome this problem we put all the extracted information (annotation results, information in the subject of the email enclosing the KID, and the data in the DEMACO system) in tabular form, and stored it into as a relational database. Then, with the support of Mastro \cite{calvanese2011mastro}, a well-known tool for semantic access to relational sources, we integrated the resources at our disposal in order to generate the ontology extensional level.

\subsection{Machine Learning-based IE}

\begin{figure*}
	\centering{
	\includegraphics[width=0.7\textwidth]{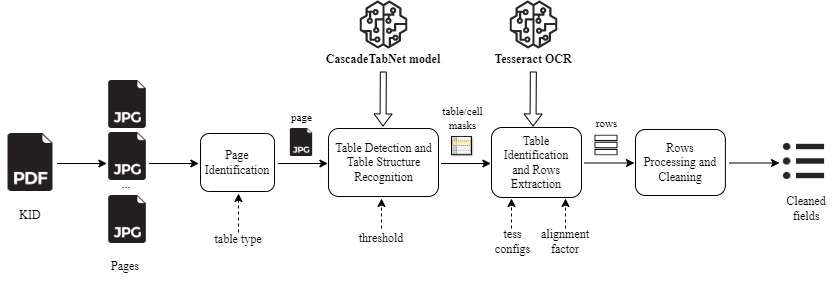}}
	\vspace*{-0.15in}
	\caption{Pipeline for table extraction}
	\label{fig:MLpipeline}
\end{figure*}

%This part of the IE 
We used ML techniques to properly deal with tabular data, and in particular to extract data from the following tables: % and fields:
\begin{itemize}
	\setlength\itemsep{0em}
	\item \textbf{Performance scenarios:}
%	,under the section "What are the risks and the potential yield?". 
It shows the potential refund and the average yield per year in all the possible scenarios: stress, unfavorable, moderate, favorable, each at fixed time periods (initial, intermediate, recommended detention).
	\item \textbf{Costs evolution:} 
%	in time, under the section "What are the costs?". 
It includes data about total costs and reduction in yield (RIY) in percentage at the same time periods considered for the performance scenarios.
	\item \textbf{Costs composition:}
%	, under the section "What are the costs?". 
	It describes the RIY for different costs categories and subcategories: input (una tantum costs), output (una tantum costs), wallet transactions (recurrent costs), others (recurrent costs), performance fees (accessory charges), overperformance fees (accessory charges). 
\end{itemize}
 
%description of the approach
We modeled the problem as a Computer Vision task, converting each page of the KID into of images. In such a setting, the Extraction involves a Table Detection task, aiming at detecting bordered or borderless/semi-bordered tables, a Table Structure Recognition task, able to identify the cells belonging to the detected table, and a Text Extraction task, to get the textual information contained inside each cell.
\\
We propose an automatic, learning-based approach, exploiting the power of Deep Learning and Artificial Neural Networks (ANN) for image processing. In particular, for subtasks 1-2 we use CascadeTabNet~\cite{cascadetabnet2020}, a recent implementation of a ANN model that benefits from convolutional layers to learn image features at different level of abstractions, detecting table areas and related cells. The model, that has proved to achieve the state-of-the-art for the considered task, combines two existing CNNs for object detection and image segmentation with a 2-stage transfer learning and image augmentation techniques. In details, HRNet is used as a feature extractor, while Cascade R-CNN predicts 3 classes of objects: (i) bordered table, (ii) borderless table and (iii) cell. In order to adapt the object detection capability of the network to the desired domain (tables inside images), the combined network, initially pre-trained on ImageNet dataset, is retrained on images containing tabular data; the transfer learning is applied firstly on a large, general dataset, allowing the network to learn table masks, and then on a smaller dataset. Depending on the specific dataset used for the transfer learning, different models are developed; we use the \textit{Modern Table Structure Recognition} model, trained on ICDAR-19.
\\
For subtask 3 instead, we use Tesseract, a popular Optical Character Recognition (OCR) engine, originally developed at HP and later by Google. The original Tesseract engine (’legacy engine’) was based on multiple steps such as page segmentation, outline and blob detection, textlines breaking, 2-pass word recognition with a trained classifier.  From Tesseract version 4, an improved engine is available, fully based on LSTM neural networks, and many new languages are supported.
\\ 
A complete view of the pipeline of the proposed approach is shown in Figure~\ref{fig:MLpipeline} and its module are:
\begin{itemize}
	\setlength\itemsep{0em}
	\item A \textbf{Page Identification Module} uses a set of strings for which we should find a match in one of the pages. For each table type (i.e. performance scenarios, costs evolution, costs composition) we select a different set of strings that identify the precise section containing the table we aim to extract.
	\item The \textbf{Table Detection and Table Structure Recognition Module} uses the pre-trained CascadeTabNet model to identify the masks (i.e. bounding boxes) of borderless/bordered tables and cells, with a certain level of confidence. The masks are returned if and only if their level of confidence is higher than the input threshold, that we experimentally tuned to 0.6. Cells are then assigned to the correspondent table depending on their positions.
	\item The \textbf{Table Identification and Row Extraction Module} applies the OCR to each cell bounding box, configuring Tesseract with the LSTM engine, the english and italian languages, and assuming a single uniform block of text for page segmentation. Again, a string match identifies the table of interest, since many tables may be present in a single page. If the table is founded, the algorithm returns a set of rows. To assign each cell to its row we use the top coordinate of the cell mask but, since very often the bounding boxes of cells belonging to the same row are not precisely aligned (i.e. one top value is slightly different than the other), we defined an alignment factor such that two cells belong to the same row if their top value differs from at most the alignment factor.
	\item Rows are arranged into fields in the \textbf{Row Processing and Cleaning Module}, assigning the numerical cells to their field type. A final Cleaning step, fully regex-based, deals with missing punctuations, extra space removal, OCR error correction and, mostly important, standardizes the numerical formats: many countries, indeed, use a comma to separate the integer from the decimal part, while others use a point. Currency units of measure are also removed from the costs, extracting just the numerical information.
\end{itemize}

The effectiveness of the results often depends on the table template: there are few tables that the algorithm fails to recognize at all, or for which some cells are not detected (mostly the table headers). Table/cell detection is indeed a challenging task, due to different reasons. Firstly, the heterogeneity of table templates adopted by the PRIIP authors makes the task harder for the neural network; templates are slightly different from the ones in ICDAR-19, because they are modern, often semi-bordered or borderless, semi-coloured, with multi-line cells. The same table type may present different number of rows or columns, just because the PRIIP type requires it, or because of a different arrangement of the information in the table (for instance, a single cell contains both a cost and a RIY). Moreover, we noticed issues in identifying multi-line cells: few times true multi-line cells are detected as two distinct cells or, on the contrary, distinct cells that appear one under the other, probably very close, are detected as a single multi-line cell. We addressed this last case in the Rows Cleaning Module: using our knowledge of the table structures, knowing which field types appear one under the other, we split the wrongly multi-line cells and assign the data to the proper field.
Regarding cell masks, we noticed also that, in few cases, the detected bounding box was quite inaccurate, cropping the text and consequently causing troubles in OCR. In order to avoid it, we consider an enlargement factor such that we run the OCR on a slightly larger area. The correctness of the results indeed, depends both on the CascadeTabNet and OCR performances. Our first results showed few inaccuracies in recognizing some characters; as an example, many '7' were recognized as '/', or some digits were recognized as alphabetic characters. Changing the OCR engine type from LSTM to legacy gave no improvement at all.  Running experiments on the documents that caused the trouble, we discovered that they mostly share the same table type (performance scenarios), the same author and, therefore, the same template, that consisted of a coloured and saturated background with a white text. Since the latest versions of Tesseract has some troubles in detecting white text on a darker background, we addressed this issue by improving Image Pre-Processing. In particular, for each cell, we checked if it has white text on a coloured background and if so we inverted the image. An important improvement has been obtained moving in this direction, removing more than half of the OCR errors in the troubling documents; the remaining errors are all of the same type ('7' as '/'). Grayscaling and adding more contrast to the image allowed to further reduce the errors.

%dpi
%A further experiment has been performed to improve the effectiveness of the results: we analyzed the impact of the image resolution, measured as dots per inch (dpi), on both the Table/Cell Detection and the OCR. Differently from what we expected, increasing significantly the dpi (from 700 to 1400) did not improve at all the accuracy; on the contrary it showed less capability of detecting cells and a larger OCR-error rate. On the other hand, very low dpis ($<$ 300) have a smaller OCR-error rate but still are less capable of detecting many cells. Experimental results prove that the optimal dpi for our data is about 300 or 400.

%% file: 04-Results/results.tex
\section{Results}

Table~\ref{tab:datasetinfo} provides some information regarding the two datasets on which we tested our approach.

\begin{table}[H]
\footnotesize
\center
\begin{tabular}{|c|c|c|c|}
\hline
 \textbf{Dataset} & \textbf{Total KIDs} & \textbf{Size} & \textbf{Manufacturers}  \\ \hline
 DATASET-1 & 1240  & ~250MB & 36 \\ \hline
 DATASET-2 & 7736 & ~3GB & 52 \\ \hline
\end{tabular}
\vspace*{-0.1in}
\caption{Dataset Info}
\label{tab:datasetinfo}
\end{table}
\vspace*{-0.1in}
Dataset-1 was created specifically as a test scenario. It is representative of document heterogeneity and includes KIDs from the years 2018-2020. Dataset-2 was selected %completely 
randomly from a sample of KIDs from the first half of 2021.

Table~\ref{tab:resrule} reports an average of precision and recall over all information extracted through the rule-based approach.

\begin{table}[H]
\vspace*{-0.15in}
\footnotesize
\center
\begin{tabular}{|c|c|c|c|}
\hline
 \textbf{Dataset} & \textbf{Precision} & \textbf{Recall} & \textbf{F-Measure} \\ \hline
 DATASET-1 & ~98\%  & ~96\% & ~97\% \\ \hline
 DATASET-2 & ~98\% & ~94\% & ~96\% \\ \hline
\end{tabular}
\caption{Results rule-based approach}
\label{tab:resrule}
\vspace*{-0.15in}
\end{table}

Table~\ref{Tab:MLresults} shows the number of tables extracted through  our ML-based approach for each table type.

%\begin{table*}[t]
%	\small
%	\center
%	\begin{tabular}{ |c|c||c|c||c|c||c|c| }
% 		\hline
%  		\multicolumn{1}{|c|}{}  & \multicolumn{2}{|c||}{\textbf{Performance Scenarios}}&\multicolumn{2}{|c||}{\textbf{Costs Evolution}} & \multicolumn{2}{|c|}{\textbf{Costs Composition}}\\
% 		\hline
% 		\textbf{Dataset} & \textbf{Extracted} & \textbf{Missing} & \textbf{Extracted} & \textbf{Missing} & \textbf{Extracted} & \textbf{Missing} \\
%		\hline
%		DATASET-1 & 1218 & 22 & 752 & 488 & 1153 & 87\\
% 		DATASET-2 &  6435 & 1301 & 3925 & 3811 & 6886 & 850\\
%		\hline
%	\end{tabular}
%	\caption{Results of the Extractor for the different table types.}
%	\label{Tab:MLresults}
%\end{table*}

\begin{table}[H]
\center
\footnotesize
\vspace*{-0.15in}
\begin{tabular}{|lc|cc|}
\hline
\multicolumn{2}{|l|}{\multirow{2}{*}{}}                                                   & \multicolumn{2}{c|}{\textbf{Dataset}}     \\ \cline{3-4} 
\multicolumn{2}{|l|}{}                                                                    & \multicolumn{1}{c|}{DATASET-1} & DATASET-2 \\ \hline
\multicolumn{1}{|c|}{\multirow{2}{*}{\vtop{\hbox{\strut \textbf{Performance}}\hbox{\strut \textbf{Scenario}}}}} & \textbf{Extracted} & \multicolumn{1}{c|}{1218}      & 6435     \\ \cline{2-4} 
\multicolumn{1}{|l|}{}                                               & \textbf{Missing}   & \multicolumn{1}{c|}{22}        & 1301     \\ \hline
\multicolumn{1}{|l|}{\multirow{2}{*}{\vtop{\hbox{\strut \textbf{Cost}}\hbox{\strut \textbf{Evolution}}}}}       & \textbf{Extracted} & \multicolumn{1}{c|}{752}       & 3925      \\ \cline{2-4} 
\multicolumn{1}{|l|}{}                                               & \textbf{Missing}   & \multicolumn{1}{c|}{488}      & 3811     \\ \hline
\multicolumn{1}{|l|}{\multirow{2}{*}{\vtop{\hbox{\strut \textbf{Cost}}\hbox{\strut \textbf{Composition}}}}}    & \textbf{Extracted} & \multicolumn{1}{c|}{1153}      & 6886       \\ \cline{2-4} 
\multicolumn{1}{|l|}{}                                               & \textbf{Missing}   & \multicolumn{1}{c|}{87}      & 850      \\ \hline
\end{tabular}
\caption{Results of the Extractor for the different table types.}
\label{Tab:MLresults}
\vspace*{-0.15in}
\end{table}

We obtained the worst recall for the Costs Evolution table; this is probably due to the fact that the extraction of this table mostly relies on just two header cells in the Table Identification step, therefore, if the model fails for any reason in detecting just those two cell masks, the table is not extracted even though the numerical cell masks are often correctly detected. This issue is less critical for the other table types, since more cells can be used for Table Identification.

%% file: 05-Conclusion/conclusion.tex
\section{Conclusion}
We are currently working to complete IE from KIDs, by attacking some of their portions that have not been so far involved in the extraction process. The next project steps concern with post-extraction data analysis, aimed to identify anomalous documents, correct errors and missing mandatory data, discover interesting correlations among data. 

%We are also investigating how to link collected data to a domain ontology, to exploit reasoning services and enhance the entire data acquisition  process (see, e.g., \cite{scafoglieri2021boosting}).